\documentclass{article}

\PassOptionsToPackage{numbers,compress}{natbib}
\usepackage[preprint]{neurips_2026}

\usepackage{iftex}
\ifPDFTeX
  \usepackage[utf8]{inputenc}
  \usepackage[T1]{fontenc}
\else
  \usepackage{fontspec}
  \usepackage{xeCJK}
\fi
\usepackage{hyperref}
\usepackage{booktabs}
\usepackage{amsfonts}
\usepackage{nicefrac}
\usepackage{microtype}
\usepackage[table]{xcolor}
\usepackage{amsmath}
\usepackage{algorithm}
\usepackage{algorithmic}
\usepackage{enumitem}
\usepackage{graphicx}
\usepackage{subcaption}
\usepackage{wrapfig}
\usepackage{multirow}
\usepackage{makecell}
\usepackage{colortbl}
\definecolor{tabgray}{gray}{0.92}

\title{Chorus II: Cross-Request Sparsity Reuse for Efficient Image-to-Video Generation}

\hypersetup{
  pdftitle={Chorus II: Cross-Request Sparsity Reuse for Efficient Image-to-Video Generation},
  pdfauthor={Hao Liu},
  pdfkeywords={image-to-video generation, sparse attention, cross-request reuse, cache acceleration},
}

\author{}
\begin{document}
\maketitle
\vspace{-25pt}
\vspace{-35pt}
\begin{center}
{\large
Hao Liu$^{1,2}$ \quad
Chenghuan Huang$^{2}$ \quad
Hao Liu$^{3}$ \quad
Xing Cai$^{3}$ \quad
Chen Li$^{3}$ \\
Ziyang Ma$^{2}$ \quad
Jing Lyu$^{3}$ \quad
Nong Xiao$^{1}$ \quad
Jiangsu Du$^{1}$ 
\par}
\vspace{-0.2em}
{\normalsize $^{1}$Sun Yat-sen University\quad $^{2}$ WeChat HPC, Tencent Inc. \quad $^{3}$ WeChat Vision, Tencent Inc. \par}
\vspace{-0.2em}
{\normalsize \texttt{\texttt{\{liuh393\}@mail2.sysu.edu.cn} \quad \texttt{\{xiaon6,dujiangsu\}@mail.sysu.edu.cn}  \quad   \{bighhliu,caderhuang,leweshaoliu,yolocai,chaselli,ziyangma,eckolv\}@tencent.com}\par}
\end{center}
\vspace{-1.3em}
\vspace{-2pt}

\begin{figure}[H]\centering
\includegraphics[width=0.95\linewidth]{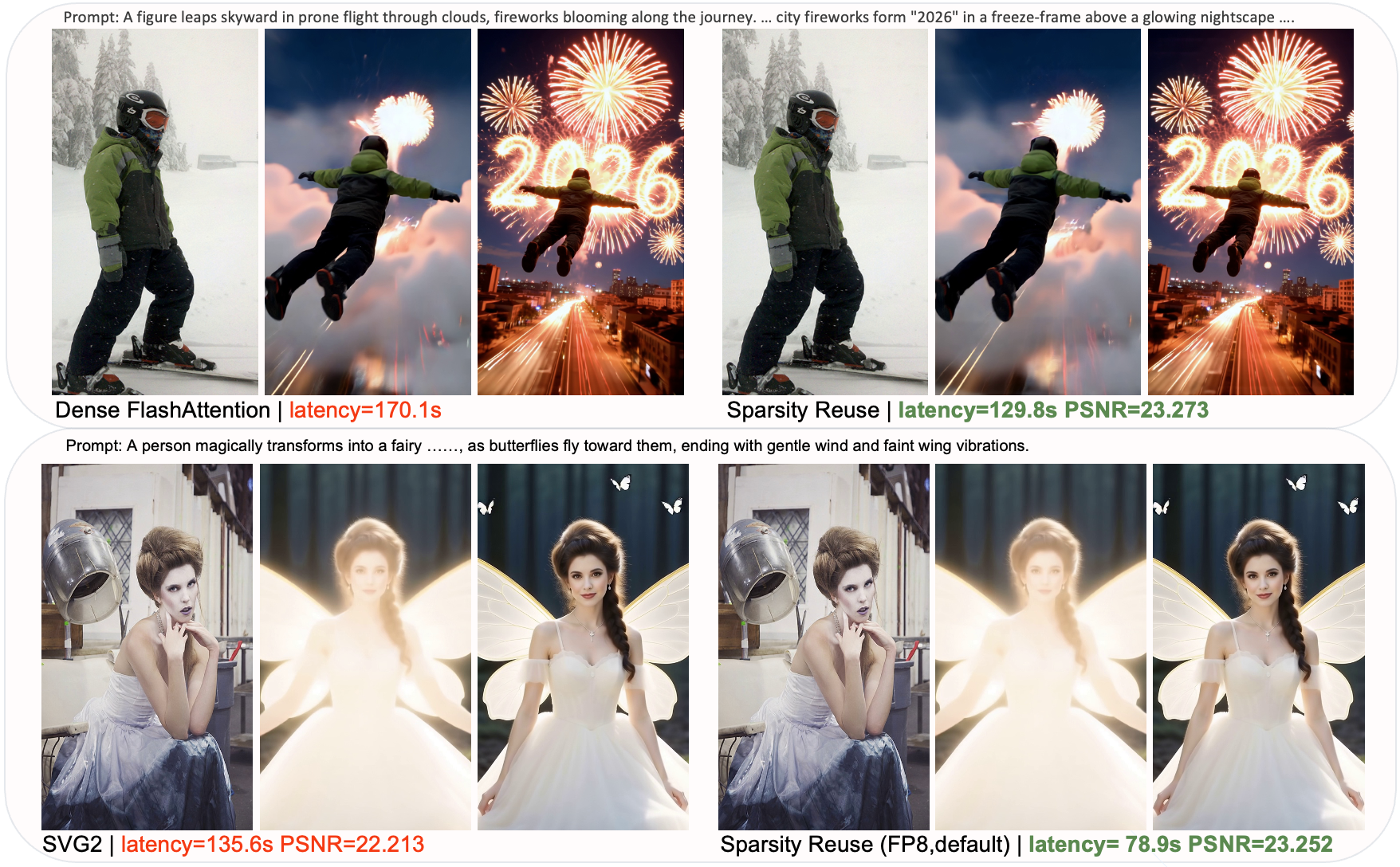}
\vspace{-3pt}
\caption{\textbf{Chorus II} accelerates video generation through cross-request sparsity reuse while maintaining high quality. On a single H20, for \textbf{4-step distilled Wan 2.2}, our default sparsity-reuse configuration achieves a 1.31$\times$ speedup in FP16 and \textbf{2.16$\times$} with our FP8 sparse-attention backend.}
\label{fig:visual_result}
\end{figure}
\vspace{-7pt}
\begin{abstract}
\vspace{-3pt}
Serving diffusion models for image-to-video generation is computationally expensive, posing significant challenges for large-scale deployment.
Real I2V workloads often contain similar requests, such as repeated effect templates, related subjects, and recurring shot layouts.
Existing cross-request acceleration methods mainly exploit this redundancy through feature reuse.
We observe that similar I2V requests also share highly consistent sparse attention patterns, enabling historical sparse masks to serve as request-conditioned priors with almost no online mask-prediction overhead.
We propose a cross-request reuse framework centered on \textbf{sparsity reuse}, with \textbf{feature reuse} as an optional extension safeguarded by a lightweight \textbf{guidance enhancement}.
Our sparsity reuse is implemented as shared sparse mask reuse, which reuses high-quality sparse masks from similar historical requests to avoid per-request online mask prediction.
Optional feature reuse applies downsampled computation to highly redundant spatiotemporal regions, mitigating boundary artifacts while preserving efficiency gains.
Guidance enhancement reinforces image/text conditioning after reuse, mitigating semantic drift and condition-adherence issues.
Experiments show that default sparsity reuse configuration preserves generation quality with a \textbf{2.16$\times$} speedup.

\end{abstract}

\section{Introduction}
Diffusion Transformers (DiTs) have demonstrated significant efficacy in generative tasks, particularly excelling in generating high-quality videos~\cite{kong2024hunyuanvideo,wan2025,zheng2024opensorademocratizingefficientvideo}, but high-resolution video generation remains expensive for online serving.
Each request requires multiple denoising steps over long latent token sequences, making latency and GPU cost the main obstacles to deployment.
To make video serving more practical, step-distillation techniques have become increasingly mainstream: by reducing the sampling steps from around 50 to fewer than 10, a step-distilled model can reduce the generation time of a 5-second video from 16.9 minutes to only a few minutes on a single NVIDIA A100 GPU (e.g., Wan2.1).

However, many acceleration methods that perform well under the 50-step setting become less effective once the steps are heavily distilled.
Dynamic sparse attention methods such as SVG2~\cite{yang2026sparse} rely on step-wise mask similarity to amortize expensive semantic clustering and mask prediction over a long denoising trajectory.
However, in the few-step regime, this overhead can no longer be sufficiently amortized and thus erodes the speedup.
Dynamic feature-cache methods such as TaylorSeer~\cite{liu2025taylorseers} predict the denoising trajectory with a Taylor expansion, yet their effectiveness drops when step-to-step similarity is low and the trajectory becomes harder to model.
Overall, these methods rely primarily on redundancy within a single request, either across sparse masks or along the denoising trajectory.
This motivates looking beyond a single request and exploiting redundancy across different requests, especially in the few-step regime where within-request redundancy is reduced and per-request overhead is harder to amortize.

\begin{wrapfigure}{r}{0.40\linewidth}
\centering
\vspace{-5pt}
\includegraphics[width=\linewidth]{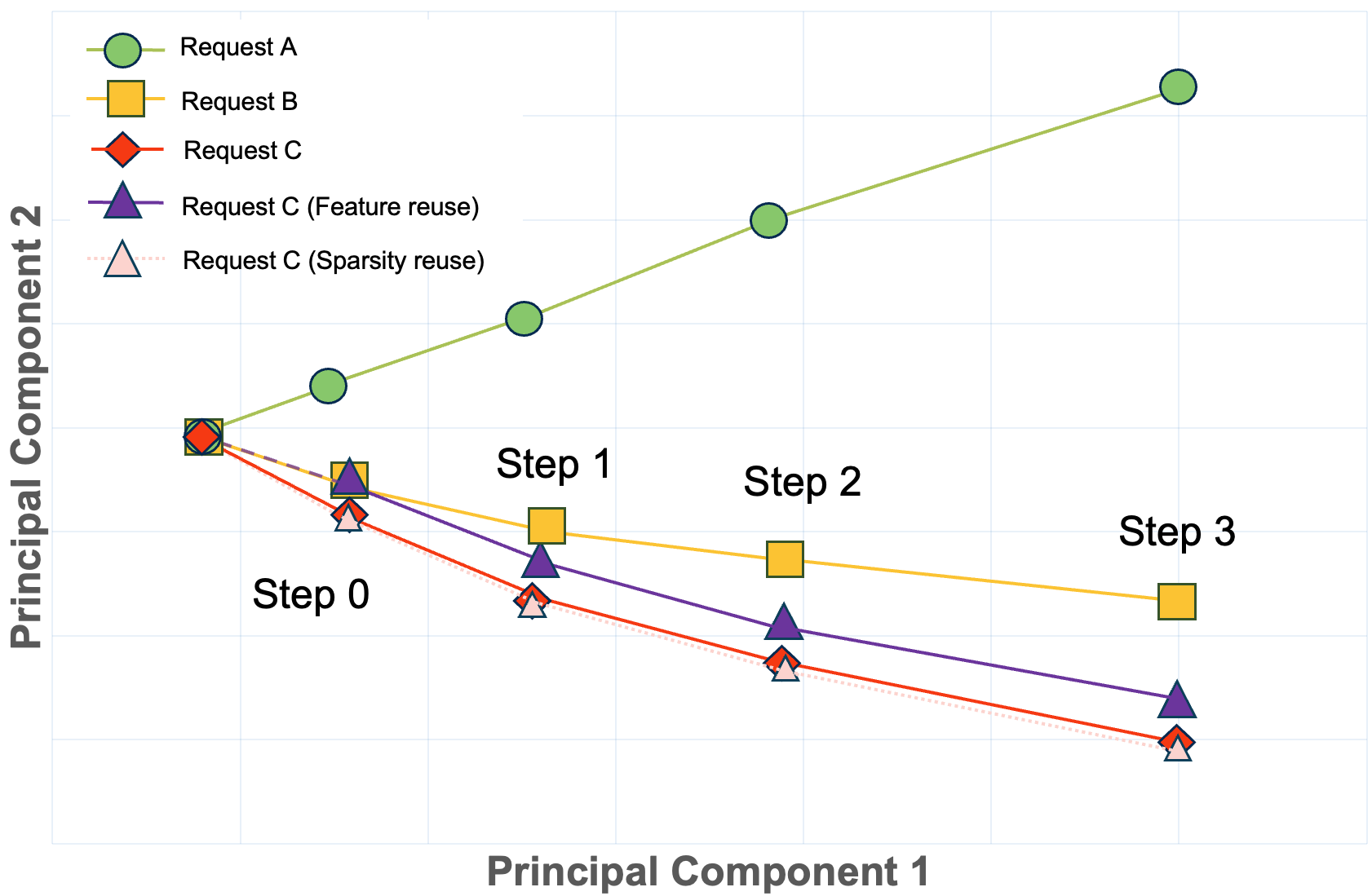}
\caption{PCA projection of feature trajectories. Requests B and C have similar prompt and image conditions.}
\vspace{-5pt}
\label{fig:feature_trajectory}
\end{wrapfigure}

Real I2V services often contain many structurally or semantically similar requests, such as recurring effect templates, the same or similar subjects, repeated motions, and related camera layouts.
Such requests share a \emph{cross-request} redundancy that intra-request acceleration methods cannot reach.
Some cross-request cache systems such as Chorus~\cite{chorus2026} and NIRVANA~\cite{agarwal2024nirvana} show that latent features can be reused across similar requests.
However, directly reusing latent features can introduce source-induced content bias, causing the resulting feature trajectory to deviate from the no-reuse baseline even when the features appear distributionally close (Fig.~\ref{fig:feature_trajectory}).
While this bias is tolerable in general scenarios, it becomes limiting in high-fidelity applications that demand minimal deviation from the no-reuse baseline. 
This calls for a more structural reuse target: sparse attention patterns, which steer computation without importing content from the source request and thus remain faithful to that baseline.
We adopt sparsity reuse as our default path, while keeping latent feature reuse as an optional and more aggressive acceleration strategy.

We found that similar I2V requests exhibit highly similar \emph{block-level sparse attention patterns}.
In our experiments, for similar requests (i.e., two I2V requests whose image-condition DINO~\cite{oquab2023dinov2} similarity exceeds 0.4), the top-$p$ (90\%) sparse masks can reach over 85\% IoU.
This suggests that the sparse mask of a historical request can serve as a request-conditioned prior for a new request, avoiding repeated online mask prediction.
Compared with static profiling, such reuse preserves request adaptivity; compared with online sparse prediction, it does not recompute the mask from scratch for every request.

This observation is particularly valuable because existing video sparse attention methods face a fundamental trade-off between mask-prediction overhead and mask accuracy.
Lightweight mask predictors are cheap but may sacrifice mask quality, whereas more accurate online construction can be so expensive that its cost cancels the benefit of sparsity.
This tension is especially acute in the few-step regime we target.
We therefore propose \emph{sparsity reuse}, implemented as \emph{shared sparse mask reuse}: high-quality sparse masks are constructed from historical requests and reused by similar future requests, enabling high-accuracy sparse attention with low mask-prediction overhead.

As an optional acceleration path beyond sparsity reuse, we further improve the latent feature reuse strategy in Chorus.
Naively skipping reusable regions can easily introduce blur, discontinuities, or tearing around region boundaries.
We therefore propose \emph{downsampled latent feature reuse}, which replaces direct skipping with lightweight downsampled context computation, allowing active tokens to access a broader receptive field while preserving the efficiency benefit of reuse.
Moreover, reuse from similar but non-identical requests may not fully align with the target image/text conditions, leading to identity inconsistency or semantic drift.
To address this issue, we design a low-cost \emph{guidance enhancement} strategy that amplifies differential conditions during attention computation and condition initialization.

Our contributions are summarized as follows:
\begin{itemize}[leftmargin=*, topsep=2pt, itemsep=2pt, parsep=0pt, partopsep=0pt]
\item We identify strong cross-request similarity in sparse attention patterns for I2V generation and propose \emph{sparsity reuse} via shared sparse mask reuse, enabling fine-grained sparse attention with near-zero online mask-prediction overhead. Experiments show that it outperforms online sparse prediction methods such as SVG2~\cite{yang2026sparse} and Sparge-Attn~\cite{zhang2025spargeattn} in the speed--quality trade-off.
\item We provide an optional speed-first extension based on cross-request \emph{downsampled latent feature reuse}, combining fine-grained spatiotemporal reuse with downsampled computation, and introduce a low-cost guidance enhancement strategy to mitigate condition inconsistency and semantic drift caused by reuse.
\item On few-step distilled I2V models, our default sparsity-reuse configuration preserves generation quality while achieving a \textbf{2.16$\times$} speedup. The optional feature reuse extension further boosts it to 2.59$\times$.
\end{itemize}

\vspace{-7pt}
\section{Related Work}
\vspace{-7pt}
\subsection{Sparse Attention for Video Generation}
Video diffusion models perform global self-attention over high-resolution spatiotemporal latents, making sparse attention an important acceleration direction.
Existing methods reduce attention cost through local windows, spatiotemporal neighborhoods, online profiling, token/block clustering, head-wise mask prediction, or step-wise propagation~\cite{xi2025sparse,yang2026sparse,zhang2025spargeattn,xia2025adaspa,shmilovich2025liteattention,ma2026haste,li2025radial,sun2025vorta}.
They mainly differ in how sparse masks are obtained: online prediction preserves input adaptivity but introduces mask-acquisition overhead, while static templates or offline calibration remove this overhead at the cost of request-level adaptivity.
In few-step distilled models, the online overhead is harder to amortize, making this trade-off more pronounced.
Our work explores a different mask-acquisition strategy: reusing high-quality sparse masks across semantically similar requests, preserving adaptivity while amortizing mask construction across requests.
\vspace{-7pt}

\subsection{Feature Cache and Reuse}
Feature reuse methods can be broadly divided into intra-request caches and cross-request reuse.
Intra-request diffusion feature caches~\cite{ma2024deepcache,lv2025fastercache,liu2025teacache,ma2025magcache,chen2024deltadit,zhao2024pab,kahatapitiya2025adacache} exploit redundancy within a single denoising trajectory by reusing, broadcasting, or predicting intermediate features across nearby timesteps.
Their effectiveness depends on strong step-to-step redundancy, which becomes weaker in few-step sampling where adjacent states are farther apart.
Cross-request, retrieval-based reuse methods~\cite{agarwal2024nirvana,chorus2026} further show that similar requests can share reusable latent information.
However, directly reusing latent features may introduce source-induced content bias, causing identity bleeding, texture contamination, or region misalignment.
Our framework retains downsampled latent feature reuse as an optional acceleration extension, while using sparse-mask reuse as the default path to avoid directly transferring latent content.

\section{Methodology}
\label{sec:method}

\subsection{Overview}
\label{sec:overview}

\begin{figure}[t]
\centering
\includegraphics[width=\linewidth]{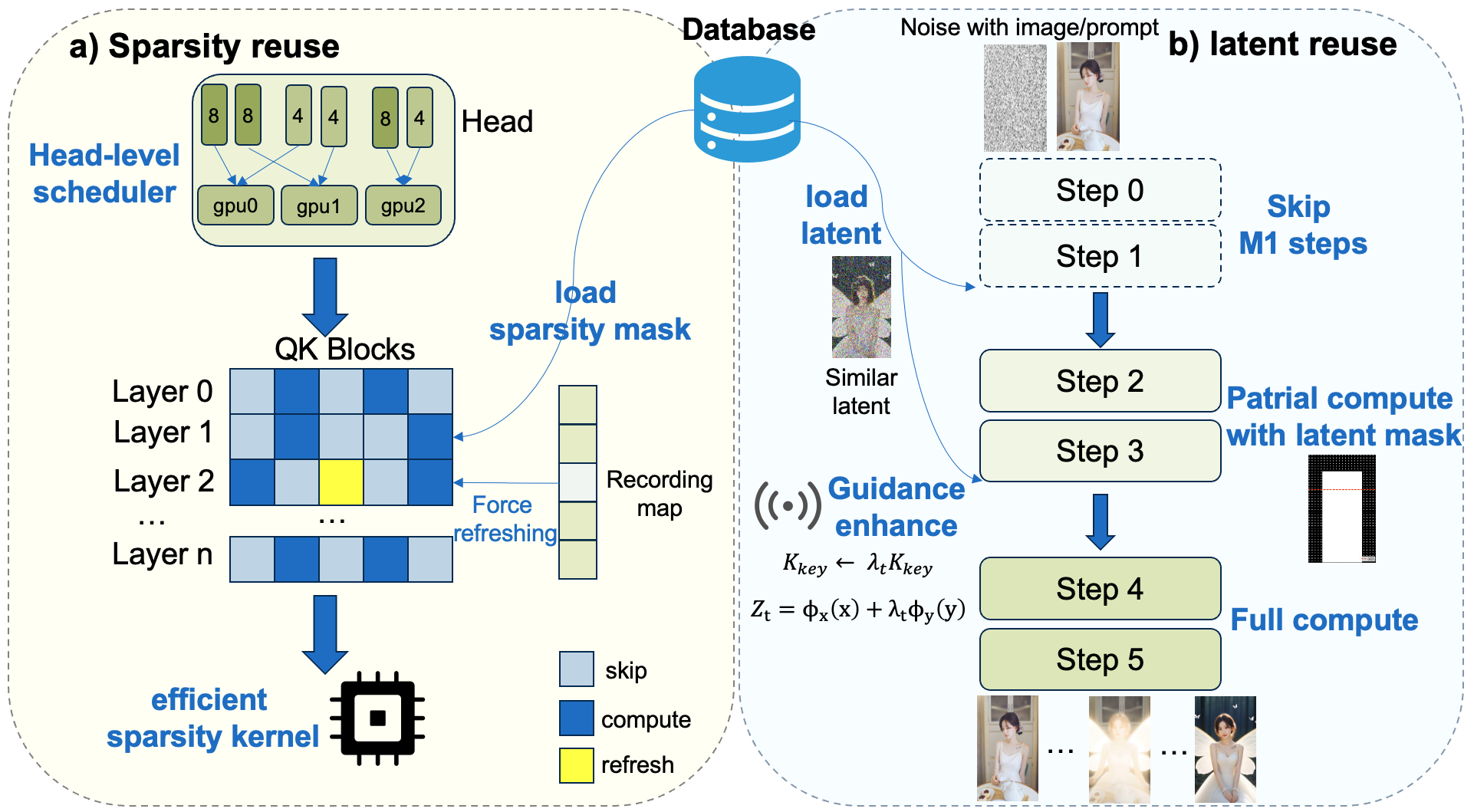}
\caption{Overview of the proposed cross-request reuse framework for efficient I2V generation.
The default path centers on reusing high-quality sparse attention masks retrieved from similar historical requests; latent feature reuse is optionally enabled for further acceleration, with lightweight guidance enhancement applied after reuse.}
\vspace{-15pt}
\label{fig:overview}
\end{figure}
\vspace{-5pt}

We propose a cross-request reuse framework for efficient I2V generation, centered on sparse-attention reuse with optional latent-feature reuse for further acceleration.
It is built on two empirical observations: \textbf{(i)} semantically similar I2V requests share global scene structure, and since block-level sparse attention patterns are governed primarily by this global structure rather than high-frequency details, such requests exhibit highly correlated sparse attention patterns; \textbf{(ii)} semantically similar requests with the same initial noise have highly similar early denoising trajectories, which makes latent feature reuse feasible.
This aligns with the coarse-to-fine nature of diffusion: at high noise levels the posterior over clean targets shares low-frequency structure, so similar requests stay largely interchangeable in early denoising---giving cross-request reuse a principled basis beyond empirical evidence.

Building on these observations, our framework has a default quality-preserving path and an optional speed-first path. 
The default path is centered on \textbf{(1)} \emph{sparsity reuse} (Sec.~\ref{sec:shared_sparse_mask}), which retrieves and reuses high-quality block-level sparse masks from similar requests. 
For more aggressive acceleration, the speed-first path additionally enables \textbf{(2)} \emph{downsampled latent feature reuse} (Sec.~\ref{sec:spatiotemporal_reuse}), which combines hierarchical region masks with downsampled context computation to convert cross-request latent feature similarity into latency savings without tearing artifacts. 
Finally, \textbf{(3)} \emph{guidance enhancement} (Sec.~\ref{sec:guidance_enhancement}) serves as a lightweight safeguard that compensates for weakened image conditioning introduced by reuse at negligible overhead.

\subsection{Sparsity Reuse}
\label{sec:shared_sparse_mask}
Video self-attention is computationally heavy.
Dynamic sparse attention predicts or constructs a sparse mask for every request, which is expensive at high accuracy, whereas static sparse masks remove mask-prediction overhead but cannot adapt to different inputs.
We observe that semantically similar I2V requests share their sparse attention patterns, which suggests a third option: instead of computing a sparse mask online for every request, we retrieve a mask from a similar historical request and reuse it for the current one.
This opens a new sparse-reuse dimension---\emph{cross-request sparse-mask reuse}---distinct from offline calibration, where all inputs share one static template, and from online dynamic sparse attention, which predicts a sparse mask from scratch.

\begin{figure}[t]
\centering
\vspace{-5pt}
\includegraphics[width=0.9\linewidth]{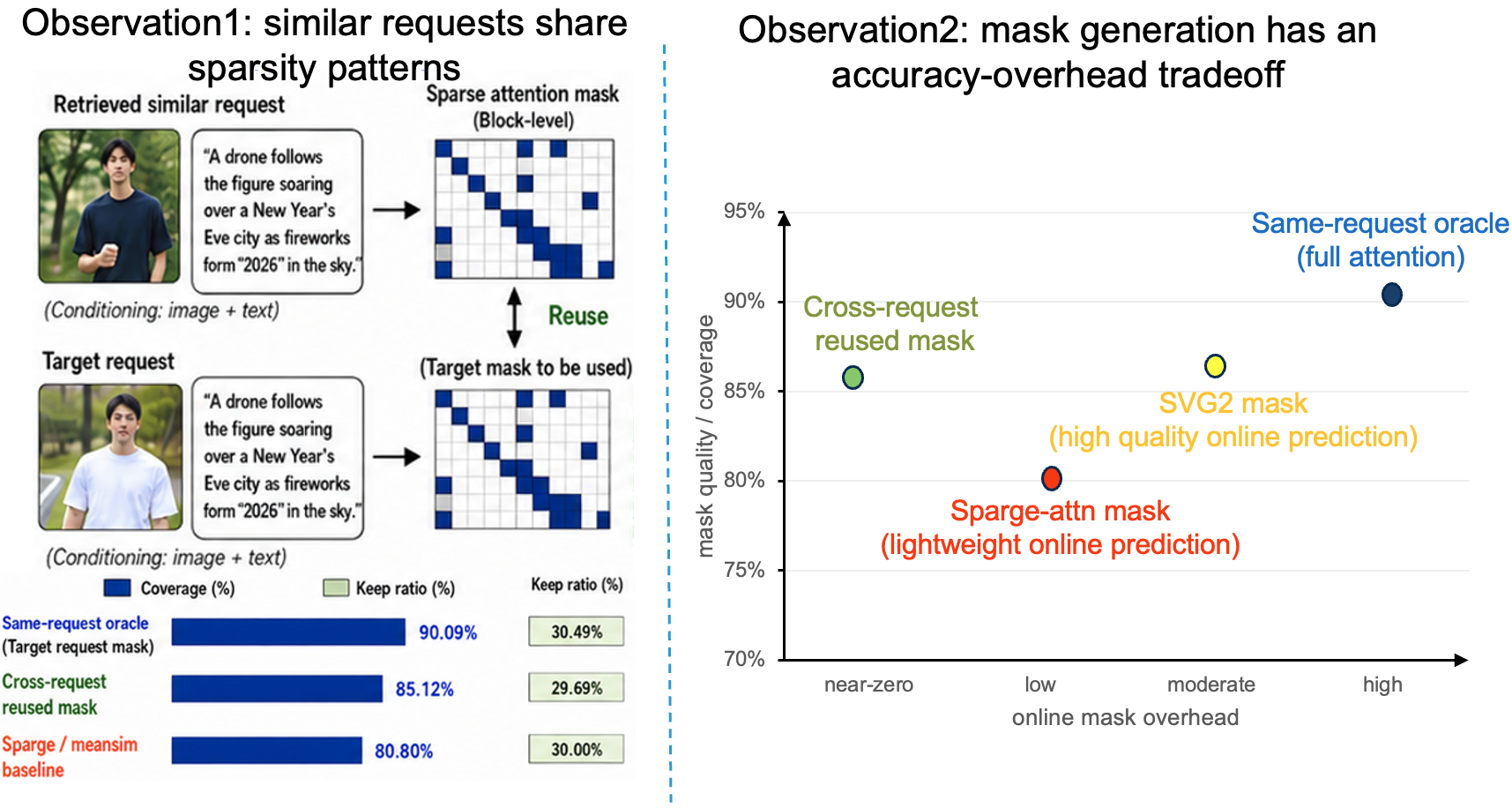}
\caption{Cross-request shared sparse mask reuse.
Sparse attention masks generated from a similar historical request can be retrieved and reused for a new I2V request, preserving most of the target's true token-level attention mass at near-zero online mask-prediction overhead.}
\vspace{-15pt}
\label{fig:shared_mask_ob}
\end{figure}

\textbf{Observation~1: sparse attention masks are reusable across similar I2V requests.}
We measure a mask by its \emph{coverage}: the fraction of the target's true token-level attention mass that falls within the blocks it selects.
The mask we study is the true-top-$p$ ($p=0.9$) mask---the ideal mask that selects blocks from the true (dense) attention until they cover a fraction $p$ of its mass.
We find that the same kind of mask, when taken from a similar request rather than the target itself and applied directly, still preserves most of the target's attention mass (Fig~\ref{fig:shared_mask_ob}).
At the same keep ratio (nearly \textbf{30\%}), it achieves \textbf{85.12\%} coverage---\textbf{only 4.97\,pp} below the same-request oracle (\textbf{90.09\%}), yet \textbf{4.32\,pp} above Sparge-Attn (\textbf{80.80\%}).
Moreover, the coverage grows monotonically as the input conditions (text and image) of the two requests become more similar, indicating that mask reusability is directly governed by cross-request similarity.

\textbf{Observation~2: existing sparse mask prediction faces an accuracy--overhead trade-off.}
Dynamic sparse attention must predict a mask before computation, yet the most accurate mask comes from a full dense pass, defeating the very purpose of acceleration. Most methods thus fall back to block-level prediction, trading accuracy for overhead.
Coarse-grained methods such as Sparge-Attn predict masks via mean-pooling followed by $QK^\top$ multiplication, but the degradation is mathematically intrinsic, as softmax and mean-pooling are not interchangeable ($\mathrm{mean}(\mathrm{softmax}(QK^\top)) \neq \mathrm{softmax}(\mathrm{mean}(Q)\,\mathrm{mean}(K)^\top)$), so a block-pair score only coarsely approximates the true attention and degrades top-$p$ mask quality.
High-accuracy methods such as SVG2 use semantic clustering and reordering for precise masks, but at a heavy prediction cost—over 30\% of the attention time in the few-step regime.
Sparse reuse elegantly recasts this accuracy--overhead dilemma as a cross-request similarity problem, amortizing accurate mask construction across similar requests instead of recomputing an approximate mask online.

\begin{wrapfigure}{r}{0.45\linewidth}
\centering
\vspace{-5pt}
\includegraphics[width=\linewidth]{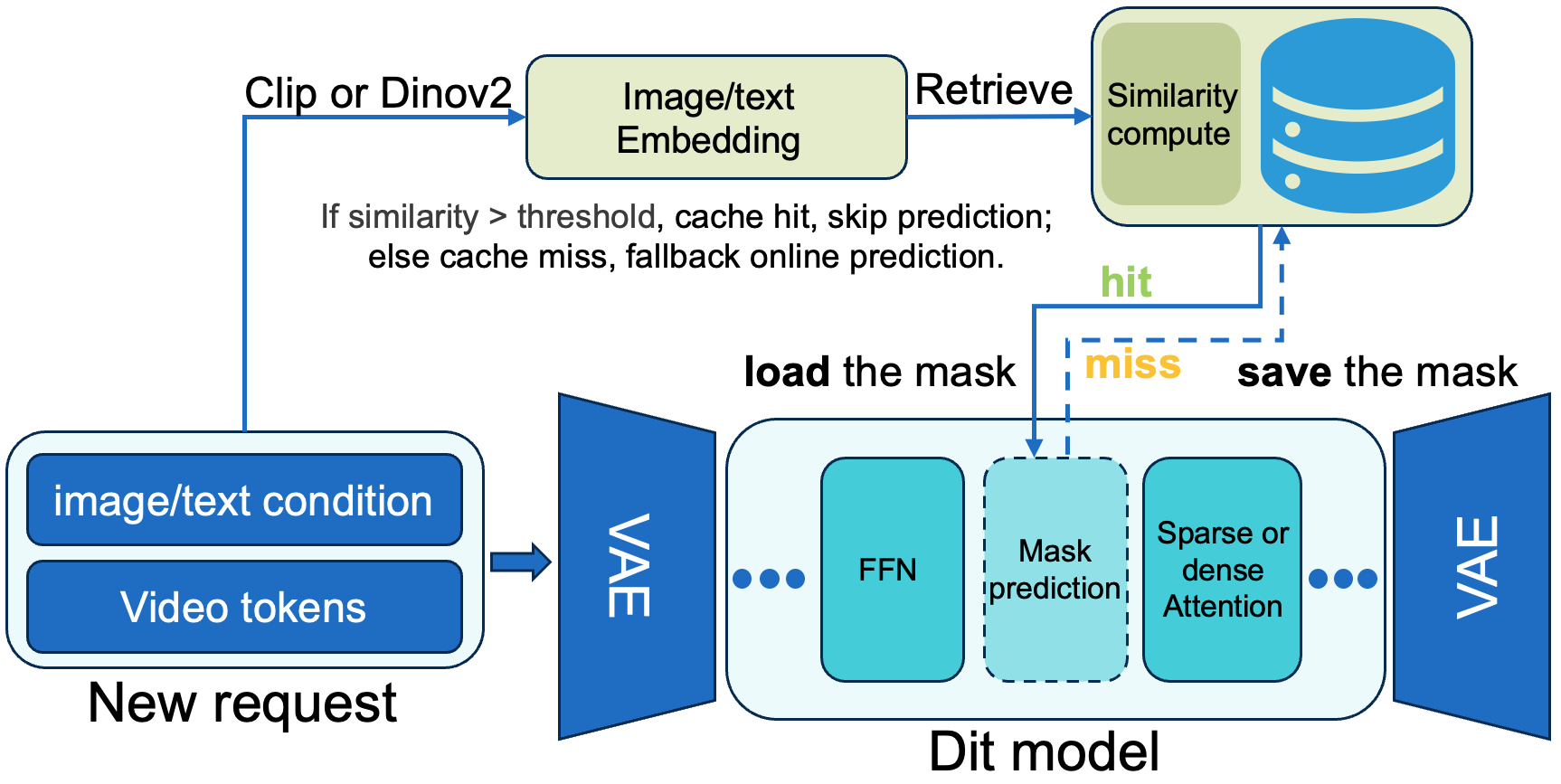}
\caption{Mask Caching and Retrieval Pipelines.}
\vspace{-5pt}
\label{fig:mask_pipeline}
\end{wrapfigure}
\textbf{Mask Caching and Retrieval.}
Motivated by Observation~1 and Observation~2, we introduce a structured \emph{mask caching and retrieval} scheme for efficient cross-request sparse mask reuse.
We treat sparse masks as cacheable request-level artifacts.
For each historical request, the system performs either dense attention or high-precision sparse attention: in the dense case we select a block-level mask from its attention map via top-$p$ with a top-$k$ floor, while in the sparse case the block-level mask is already produced and is reused directly. The resulting high-quality sparse masks are then stored, indexed by denoising step, transformer layer, and attention head.
Concretely, we encode the text prompt and the conditioning image of each request into embeddings and store them in a vector database (e.g., ChromaDB). 
For a new I2V request, we perform a $k$-nearest-neighbor search under cosine similarity over the joint text and image embeddings; if the similarity to the top neighbor exceeds a threshold $\tau$, its cached sparse masks are loaded and used directly by the current block sparse attention.
Because the sparse pattern is consistent across similar requests, the reused mask still covers most of the target's true attention mass. Moreover, since the historical request has to perform its own attention computation during generation anyway, the high-quality sparse mask is recorded as a by-product of this computation and incurs virtually no extra overhead, while the new request only performs a cache lookup.
If no sufficiently similar request is found, the system falls back to lightweight online mask prediction or dense attention to avoid mask-mismatch--induced quality drops.

\textbf{Safety fallback against information loss.}
Although similar requests share closely correlated sparsity patterns, a slight drift inevitably exists between them, and directly reusing the source mask can introduce additional attention bias.
To guarantee sufficient information flow while keeping the reuse benefit, we design a lightweight two-level safety fallback mechanism.
\textbf{(1) Block-pair visit refreshing.}
We maintain a binary visited map $V[h, q, k]\in\{0,1\}$ that records, for the current request, whether the block pair $(h, q\text{-block}, k\text{-block})$ has been activated by the sparse mask in the current layer group.
We define a layer group as $N$ consecutive transformer layers (denoted \texttt{layer\_flash}; we use $N=8$).
Within each group, the visited map is updated layer by layer; when executing the \emph{last} layer of the group, any block pair that is still unvisited is forced into the sparse mask, ensuring that every Q/K block pair is touched at least once per group.
The visited map is then reset at the start of the next group.
\textbf{(2) Minimum top-$k$ guarantee.}
We observe that some attention heads exhibit extremely peaked distributions, where the attention mass concentrates on only a few block pairs.
If the sparse mask is selected purely by a top-$p$ criterion, such heads retain only a handful of block pairs, and under cross-request reuse even a slight distributional drift between the source and target requests can misalign these few blocks and cause severe information loss.
To mitigate this, we impose an additional minimum top-$k$ constraint (denoted \texttt{min\_top\_k}) when caching the sparse mask, ensuring that every head retains at least a fraction $k_{\min}$ of its key blocks and thus preserves a safety margin against drift.
With these two safeguards, our method recovers a substantial fraction of the quality loss that pure sparse reuse would otherwise incur, at only a marginal compute overhead per request.

\textbf{Head-level scheduling for load balancing.}
Since our joint sparsity strategy (top-$p$ selection with a top-$k$ floor) assigns a different sparsity level to each attention head, the per-head workloads are imbalanced; under Ulysses sequence parallelism~\cite{jacobs2023deepspeed}, where each GPU is responsible for a subset of heads after the all-to-all exchange, this directly translates into imbalanced per-GPU loads and a straggler effect.
A key advantage of sparse reuse is that the sparsity distribution of every head is known in advance, which allows us to schedule heads across GPUs ahead of time to balance the load.
Specifically, we adopt a greedy strategy that reorders heads before the all-to-all communication, assigning heads with lower sparsity (i.e., heavier workload) to the GPUs with the lightest current load.
This scheduling achieves about 7.28\% end-to-end speedup under 8-GPU Ulysses parallelism (Table~\ref{tab:head_level}).

\textbf{Efficient block-sparse kernel.}
To translate the reused sparse masks into wall-clock speedups, we build on two block-sparse attention backends. For the non-quantized (fp16) path we use FlashInfer~\cite{ye2025flashinfer}, and our default backend is adapted from the open-source quantized backend of SpargeAttn~\cite{zhang2025spargeattn}, combining block sparsity with FP8 quantization for more aggressive acceleration.
Both backends skip pruned block pairs and evaluate only the retained ones in $QK^\top$, softmax, and $PV$; the FP8 path further uses fast low-precision matrix multiplication with high-precision accumulation at numerically sensitive stages to preserve accuracy.
At a \textbf{35.2\%} keep ratio, our default FP8 backend achieves a \textbf{6.71$\times$} speedup over FlashAttention-2.

\begin{figure}[t]
\centering
\includegraphics[width=0.85\linewidth]{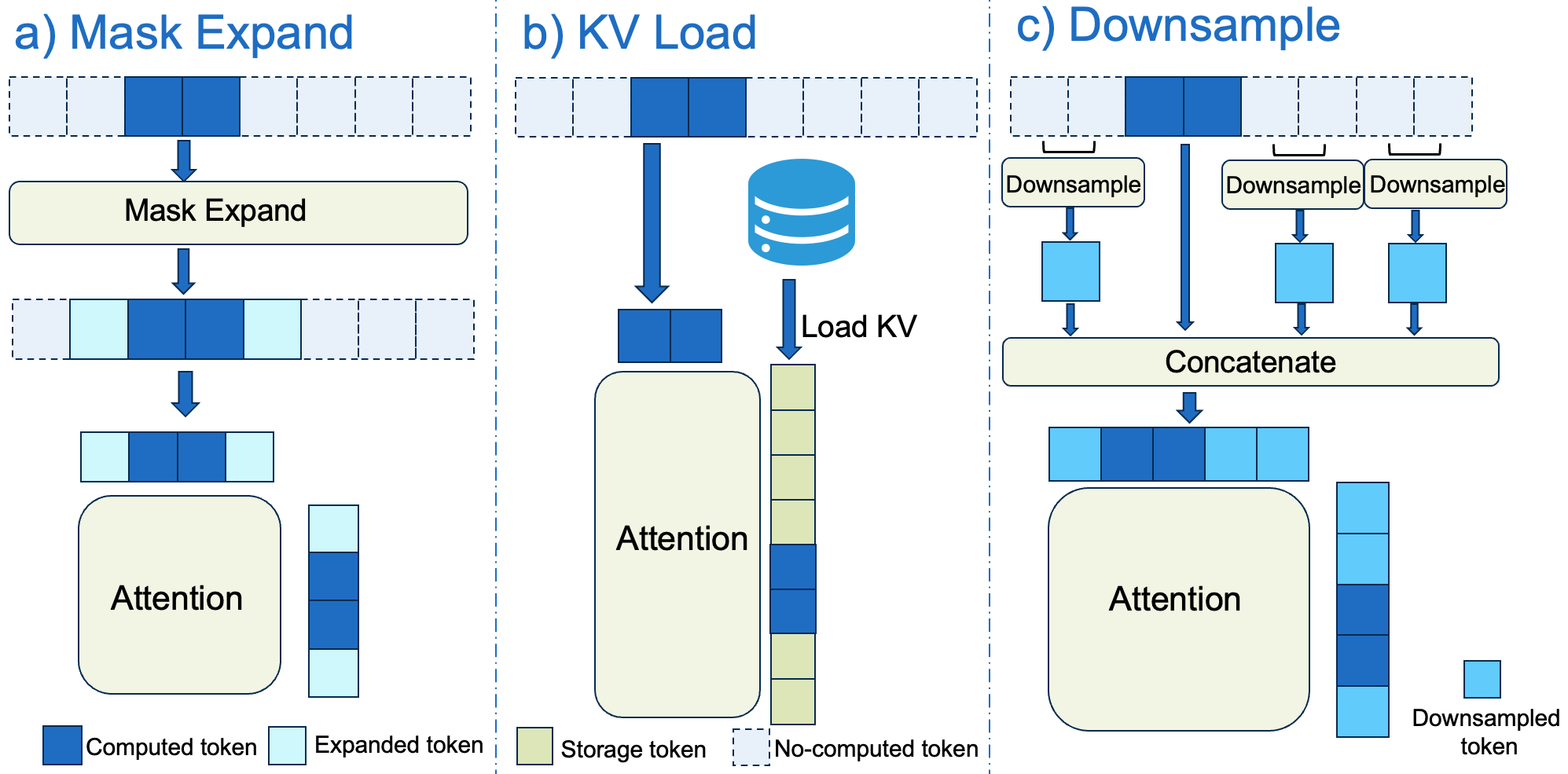}
\caption{Three approaches to alleviate the inconsistency issues introduced by fine-grained latent feature reuse: (a) mask expansion, (b) KV loading, and (c) downsampling.
}
\vspace{-15pt}
\label{fig:downsample}
\end{figure}
\vspace{-5pt}

\subsection{Downsampled Latent Feature Reuse}
\label{sec:spatiotemporal_reuse}
Building upon prior work, we start from a fine-grained latent feature reuse strategy similar to Chorus~\cite{chorus2026}, which utilizes masks to partition the video into low-similarity computed regions and high-similarity non-computed regions (reuse zones).
However, such reuse with region isolation severely blocks the information flow from the reused non-computed regions to the computed regions, inevitably triggering boundary blurring and spatial incoherence (tearing artifacts) at the region interfaces.
While Chorus attempts to mitigate this issue through expanded boundary masking (Fig.~\ref{fig:downsample}(a)), it cannot be fully resolved without global context.
An alternative approach to facilitate this cross-zone information flow is to cache and store the Key-Value (KV) states of the non-computed regions across all layers.
During the forward pass of each attention block, the cached KV states are concatenated with the active sequence's KV states, while keeping the Query ($Q$) tensor restricted to the active zone.
Although this cross-attention concatenation effectively restores context awareness and relieves boundary tearing, it introduces substantial storage and computational overheads.
For instance, in Wan2.2, caching the intermediate KV states across all layers demands approximately \textbf{160$\times$} more storage capacity than simply caching the raw latent features.

To address this limitation with minimal resource footprints, we propose \emph{downsampled latent feature reuse}, a hybrid feature-reuse method using \emph{downsampled context computation}.
Instead of completely discarding the non-computed tokens or caching their massive KV states, we apply a spatiotemporal downsampling pool directly to these non-computed regions.
The complete active sequence is then concatenated with the spatiotemporally downsampled non-computed sequence before being fed into subsequent attention layers.
This hybrid computation ensures that the active tokens retain a virtually complete, globally coherent receptive field throughout the generation process, fundamentally eliminating boundary tearing at a negligible fraction of the computational and memory cost of full-resolution modeling.

\subsection{Guidance Enhancement}
\label{sec:guidance_enhancement}

Both sparsity reuse and downsampled latent feature reuse introduce small biases that accumulate across denoising steps and weaken the image condition, causing identity drift or composition shift.
We counter this with two lightweight, training-free enhancements that merely recalibrate existing condition pathways at negligible cost ($<20$\,ms/step).

\textbf{First-frame key enhancement.}
After the QKV projection in self-attention, we scale the keys of first-frame tokens by a step-dependent factor $\beta_{\mathrm{step}}$:
\begin{equation}
K[:,0:N_0,:,:] \leftarrow \beta_{\mathrm{step}}\,K[:,0:N_0,:,:],
\end{equation}
where $N_0=H_{\mathrm{patch}}W_{\mathrm{patch}}$ is the number of first-frame tokens, leaving Q and V unchanged.
This raises the competitiveness of first-frame columns under softmax, so subsequent frames read first-frame information more thoroughly.

\textbf{Image embedding enhancement and decay schedule.}
At the patch embedding stage, we split the convolution weight along channel groups $W=[W_x \mid W_y]$ ($x$: noisy latent, $y$: image condition) and amplify the image branch:
\begin{equation}
e = \mathrm{Conv3d}(x,W_x) + \alpha_{\mathrm{step}}\,\mathrm{Conv3d}(y,W_y) + b,
\end{equation}
which equals $\mathrm{Conv3d}([x,\alpha_{\mathrm{step}} y],W)$ and needs no retraining.
Both enhancements are enabled only after reuse starts ($i \geq \mathrm{CACHE\_STEP}$) and decay along sampling steps.

\section{Experiments}
\label{sec:experiments}

\subsection{Setup}
\label{sec:settings}

\textbf{Models.}
We adopt Wan2.2-I2V as our main experimental model, an advanced open-source video generation model built on a Mixture-of-Experts (MoE) architecture.
To evaluate the acceleration potential of our method in the more practical industrial setting of few-step distillation, we apply the lightx2v LoRA~\cite{lightx2v} to reduce the sampling steps from 50 to 4.
By default, all evaluated videos are generated at a 720p resolution with 61 frames and 3600 tokens per frame, utilizing DPM-Solver++~\cite{Lu_2025} as the default diffusion solver.
All experiments and benchmark latency measurements are conducted on NVIDIA H20 GPUs.

\textbf{Dataset.}
COCO 2017 images~\citep{lin2014microsoft} are used as I2V condition images, with text prompts drawn from real-world scenarios of our video generation services.
To measure the algorithmic effect of sparsity reuse itself, all reuse variants are evaluated under a warm cache; at deployment, requests with image-condition similarity below $\tau=0.4$ fall back to online sparse attention like Sparge-Attn.

\textbf{Metrics.}
We group our evaluation metrics into two categories: video quality and efficiency.
Video quality is assessed along three dimensions: \emph{general visual quality}, \emph{condition adherence}, and \emph{fidelity to the dense baseline}.
General visual quality is summarized by VBench-Q~\cite{huang2024vbench}, computed as the average of VBench subject consistency, background consistency, motion smoothness, dynamic degree, aesthetic quality, and imaging quality.
Condition adherence is measured by image-condition adherence (DINO-I) and text-condition adherence (CLIP-T), where DINO-I follows the VBench-I2V~\cite{huang2024vbenchcomprehensiveversatilebenchmark} subject-consistency protocol (DINO feature similarity to the condition image) and CLIP-T is based on the CLIP score~\cite{hessel2021clipscore}.
Fidelity to the dense baseline is characterized by PSNR, SSIM, LPIPS, and CLIP-F, where CLIP-F denotes the CLIP frame-feature similarity between the accelerated and dense-baseline videos (distinct from the text-prompt CLIP-T above). 
Efficiency is measured by end-to-end latency and speedup.

\textbf{Baseline and Compared methods.}
We adopt the 4-step distilled model with dense FlashAttention~\cite{dao2022flashattention} as our baseline.
For comparison, we select three representative sparse-attention methods: SVG2~\cite{yang2026sparse} for high-accuracy dynamic sparsity, Sparge-Attn~\cite{zhang2025spargeattn} for coarse-grained dynamic sparsity, and Radial-Attn~\cite{li2025radial} for static sparsity.
We vary their sparsity thresholds to examine their behavior under different sparsity levels (i.e., different accuracy-efficiency trade-offs), while keeping all other settings at the defaults from their respective papers and official repositories.
For our cross-request reuse framework, we evaluate the default sparsity-reuse path as well as an variant that combines sparsity reuse with downsampled latent feature reuse.
For sparsity reuse, we fix \texttt{layer\_flash}=8, \texttt{Q\_block\_size}=64, and \texttt{K\_block\_size}=128, and report two configurations: the \emph{default} setting (\texttt{top\_p}=0.95, \texttt{min\_top\_k}=0.1) and the \emph{turbo} setting (\texttt{top\_p}=0.9, \texttt{min\_top\_k}=0.0).

\subsection{Main Results}
\label{sec:main_results}
Table~\ref{tab:main_results} reports the main comparison on Wan2.2-I2V under the 4-step setting.
We compare dense quantization, representative sparse-attention methods with different routing strategies, and our cross-request reuse variants to analyze the resulting quality--speed trade-offs.

\begin{table*}[t]
\centering
\caption{Main experimental results on Wan2.2-I2V under the 4-step setting. Our default configuration is shaded.}
\label{tab:main_results}
\small
\setlength{\tabcolsep}{4.5pt}
\renewcommand{\arraystretch}{1.15}
\resizebox{\textwidth}{!}{
\begin{tabular}{lccccccccc}
\toprule
\textbf{Method} & \textbf{VBench-Q}\,$\uparrow$ & \textbf{DINO-I}\,$\uparrow$ & \textbf{CLIP-T}\,$\uparrow$ & \textbf{PSNR}\,$\uparrow$ & \textbf{SSIM}\,$\uparrow$ & \textbf{LPIPS}\,$\downarrow$ & \textbf{CLIP-F}\,$\uparrow$ & \textbf{Latency}\,$\downarrow$ & \textbf{Speedup}\,$\uparrow$ \\
\midrule
\multicolumn{10}{l}{\emph{Dense (reference, no sparsification)}} \\
FlashAttention(baseline)            & 0.790 & 0.922 & 24.12 & --     & --     & --     & --     & 170.1\,s & 1.00$\times$ \\
SageAttention~\cite{zhang2025sageattention} & 0.789 & 0.922 & 24.05 & 24.817 & 0.852 & 0.093 & 0.9894 & 91.8\,s  & 1.85$\times$ \\
\midrule
\multicolumn{10}{l}{\emph{Sparse attention}} \\
Radial-Attn                 & 0.786 & 0.920 & 24.14 & 20.043 & 0.723 & 0.194 & 0.9761 & 130.0\,s & 1.31$\times$ \\
Sparge-Attn (top-$k$=0.08)  & 0.779 & 0.894 & 24.11 & 16.157 & 0.591 & 0.324 & 0.9499 & 68.3\,s  & 2.49$\times$ \\
Sparge-Attn (top-$k$=0.35)  & 0.788 & 0.922 & 24.10 & 20.879 & 0.741 & 0.169 & 0.9791 & 76.2\,s & 2.23$\times$ \\
Sparge-Attn (top-$k$=0.5)   & 0.788 & 0.921 & 24.11 & 21.481 & 0.772 & 0.153 & 0.9824 & 80.5\,s & 2.11$\times$ \\
SVG2 (default,top-$p$=0.95)& 0.787 & 0.917 & 24.14 & 23.124 & 0.817 & \textbf{0.118} & \textbf{0.9866} & 160.0\,s & 1.06$\times$ \\
SVG2 (fast,top-$p$=0.9)   & 0.788 & 0.921 & 24.10 & 22.213 & 0.791 & 0.135 & 0.9840 & 135.6\,s & 1.25$\times$ \\
\midrule
\multicolumn{10}{l}{\emph{Cross-request reuse (Ours)}} \\
Ours (sparsity + feature reuse) & 0.787 & \textbf{0.924} & \textbf{24.50} & 13.211 & 0.500 & 0.461 & 0.9305 & \textbf{65.7\,s} & \textbf{2.59$\times$} \\
Ours (sparsity reuse, fp16) & 0.787 & 0.918 & 24.11 & \textbf{23.273} & \textbf{0.818} & 0.119 & 0.9860 & 129.8\,s & 1.31$\times$ \\
\rowcolor{tabgray}
Ours (sparsity reuse)       & 0.787 & 0.918 & 24.16 & 23.252 & 0.818 & 0.120 & 0.9859 & 78.9\,s  & 2.16$\times$ \\
Ours (sparsity reuse, turbo)& 0.788 & 0.923 & 24.20 & 21.128 & 0.759 & 0.163 & 0.9802 & 75.6\,s & 2.25$\times$ \\
\bottomrule
\end{tabular}}
\vspace{-15pt}
\end{table*}

\textbf{Quality Evaluation.}
In terms of general visual quality, most methods under non-extreme acceleration settings maintain VBench-Q within 0.786--0.790, indicating that moderate attention acceleration does not noticeably degrade overall visual quality. The clear exception is the extremely sparse Sparge-Attn setting (top-$k$=0.08), whose VBench-Q drops to 0.779, suggesting that overly aggressive sparsification can harm generation quality.
Condition adherence shows a similar trend. For most methods, DINO-I and CLIP-T remain close to the 4-step baseline, indicating that image- and text-condition adherence is largely preserved.
In contrast, Sparge-Attn (top-$k$=0.08) reduces DINO-I to 0.894, showing weakened adherence to the input image. 
Our variant with feature reuse achieves the highest DINO-I and CLIP-T scores, suggesting that even when feature reuse changes the generation trajectory, the output can still follow the input conditions well.

The largest differences appear in fidelity to the dense baseline. Among sparse-attention methods, SVG2 (top-$p$=0.95) provides the strongest fidelity to the dense baseline, clearly outperforming coarse-grained dynamic sparsity represented by Sparge-Attn and static sparsity represented by Radial-Attn.
Our sparsity reuse achieves SVG2-level fidelity: it obtains slightly higher PSNR/SSIM (23.252/0.818 vs. 23.124/0.817) and comparable LPIPS/CLIP-F (0.120/0.9859 vs. 0.118/0.9866). This shows that cross-request sparse-mask reuse can preserve the fidelity of high-accuracy dynamic sparsity without requiring expensive online routing.
Feature reuse represents a more aggressive quality--speed trade-off. Since it reuses additional intermediate features, the generated video may deviate more from the baseline trajectory, leading to lower PSNR, SSIM, LPIPS, and CLIP-F. Nevertheless, its VBench-Q and condition-adherence scores remain strong, indicating that it may not closely reproduce the baseline video but can still generate plausible, well-conditioned outputs.

\textbf{Efficiency Evaluation.}
The efficiency results reveal distinct quality--speed regimes. High-fidelity dynamic sparsity, represented by SVG2, preserves fidelity but provides limited end-to-end acceleration due to online sparse-routing construction: the default setting reaches only 160s (1.06$\times$), and the turbo setting improves to 135.6s (1.25$\times$). Radial-Attn avoids dynamic routing with a static sparse pattern, but its acceleration is also limited at 130s (1.31$\times$).
Sparge-Attn achieves higher speed through coarse-grained, low-overhead sparsity and quantized attention kernels. With top-$k$=0.35/0.5, its latency is 76.2--80.5s; reducing top-$k$ to 0.08 further lowers latency to 68.3s (2.49$\times$), albeit at a clear quality cost

Our sparsity reuse avoids online sparse-mask search by reusing masks across requests. In fp16, it reaches 129.8s (1.31$\times$); with the quantized attention kernel, it further improves to 78.9s (2.16$\times$). This latency is comparable to Sparge-Attn (top-$k$=0.35/0.5), but with much higher fidelity to the dense baseline. Meanwhile, it matches SVG2 in fidelity to the dense baseline while reducing its latency by roughly half, demonstrating that sparse-mask reuse mitigates the routing-overhead bottleneck of high-fidelity dynamic sparsity.
The turbo sparsity-reuse configuration further reduces latency to 75.6s (2.25$\times$), and combining feature reuse achieves the lowest latency, 65.7s (2.59$\times$). As discussed in the quality evaluation, this configuration is better viewed as a speed-first operating point due to its larger fidelity drop.
Overall, our method improves the practical quality--speed Pareto frontier: it provides substantially higher fidelity to the dense baseline than Sparge-Attn at comparable latency, and achieves SVG2-level fidelity to the dense baseline with about half the latency.

\subsection{Mask Quality and Routing Overhead}
\label{sec:ablation_a}
We now look inside the attention path to isolate the mask-acquisition mechanism, comparing online sparse mask construction against cross-request mask retrieval in sparsity quality and acquisition overhead.
\emph{Routing/Retrieval Time} measures mask generation overhead, \emph{GT Coverage} the fraction of ground-truth attention mass covered by the mask, and \emph{Attention Latency} the routing overhead plus the attention kernel.
To compare routing cost across methods, we define
\begin{equation}
\label{eq:routing_overhead_ratio}
\text{Routing Overhead Ratio} = \frac{\text{Routing Time}}{\text{Attention Latency}},
\end{equation}
where a lower ratio indicates a path closer to dense attention's zero-overhead behavior.
To expose the routing--quality trade-off, we evaluate SVG2 in its default setting and in two accelerated variants that progressively reduce KMeans iterations: \emph{fast} (\texttt{iter\_init}=15, \texttt{iter\_step}=2) and \emph{turbo} (\texttt{iter\_init}=10, \texttt{iter\_step}=1).

\begin{table}[t]
\centering
\caption{Mask acquisition on the attention path: online routing vs. cross-request reuse. Reported latencies cover only the routing stage and attention kernel. \emph{Routing OH ratio} = Routing Time / Attention Latency. Ours here refers to sparsity reuse.}
\label{tab:ablation_sparse_reuse}
\small
\setlength{\tabcolsep}{3pt}
\renewcommand{\arraystretch}{1.08}
\resizebox{\linewidth}{!}{
\begin{tabular}{lcccccc}
\toprule
\textbf{Method} & \textbf{Routing time} $\downarrow$ & \textbf{Routing OH ratio} $\downarrow$ & \textbf{Density} $\downarrow$ & \textbf{GT coverage} $\uparrow$ & \textbf{Attn latency} $\downarrow$ & \textbf{Speedup} $\uparrow$ \\
\midrule
FlashAttention                      & 0\,ms     & 0.0\%   & 100.0\%     & 100.0\%   & 751\,ms   & 1.00$\times$ \\
SVG2 (default,topp=0.9)             & 339\,ms   & 54.7\%  & 29.1\%      & 90.1\% & 620\,ms   & 1.21$\times$ \\
SVG2 (fast,topp=0.9)                & 205\,ms   & 40.7\%  & 29.5\%      & 89.5\% & 504\,ms   & 1.49$\times$ \\
SVG2 (Turbo,topp=0.9)               & 145\,ms   & 31.3\% & 32.7\%      & 89.2\% & 464\,ms   & 1.62$\times$ \\
Ours (fp16,default)                 & 0\,ms     & 0.0\%   & 39.5\%      & 92.0\% & 321\,ms   & 2.34$\times$ \\
\midrule
SageAttention                       & 0\,ms     & 0.0\%   & 100.0\%     & 100.0\%   & 252\,ms   & 2.98$\times$ \\
Sparge-Attn (fp8,topk=0.4)          & 6.3\,ms   & 5.2\%   & 40.0\%      & 86.6\% & 121\,ms & 6.21$\times$ \\
Ours (fp8,turbo)                    & 0\,ms     & 0.0\%   & 35.2\%      & 88.1\% & 112\,ms & 6.71$\times$ \\
\bottomrule
\end{tabular}}
\vspace{-15pt}
\end{table}

The results separate two regimes of online routing.
High-cost routing (SVG2) attains the most accurate masks among online methods (highest GT coverage $\sim\!90\%$), but its routing stage dominates the attention path with a Routing Overhead Ratio of $31$--$55\%$, eroding most of the sparsity gain---primarily due to \texttt{flashinfer.plan} under irregular block sizes and the costly KMeans initialization that cannot be amortized in the few-step setting.
Fast routing (Sparge-Attn) predicts masks cheaply but at lower coverage ($86.6\%$).
Cross-request reuse breaks this trade-off: by retrieving high-quality historical masks, Ours incurs \emph{zero} routing overhead, with the \emph{default} version even exceeding the best SVG2 in coverage ($92.0\%$ vs.\ $90.1\%$) and the \emph{turbo} version staying above Sparge-Attn ($88.1\%$ vs.\ $86.6\%$). It thus Pareto-dominates online routing, achieving the lowest attention latency in each precision regime ($321$\,ms in fp16, $112$\,ms in fp8).

\begin{wrapfigure}{r}{0.45\linewidth}
\centering
\includegraphics[width=\linewidth]{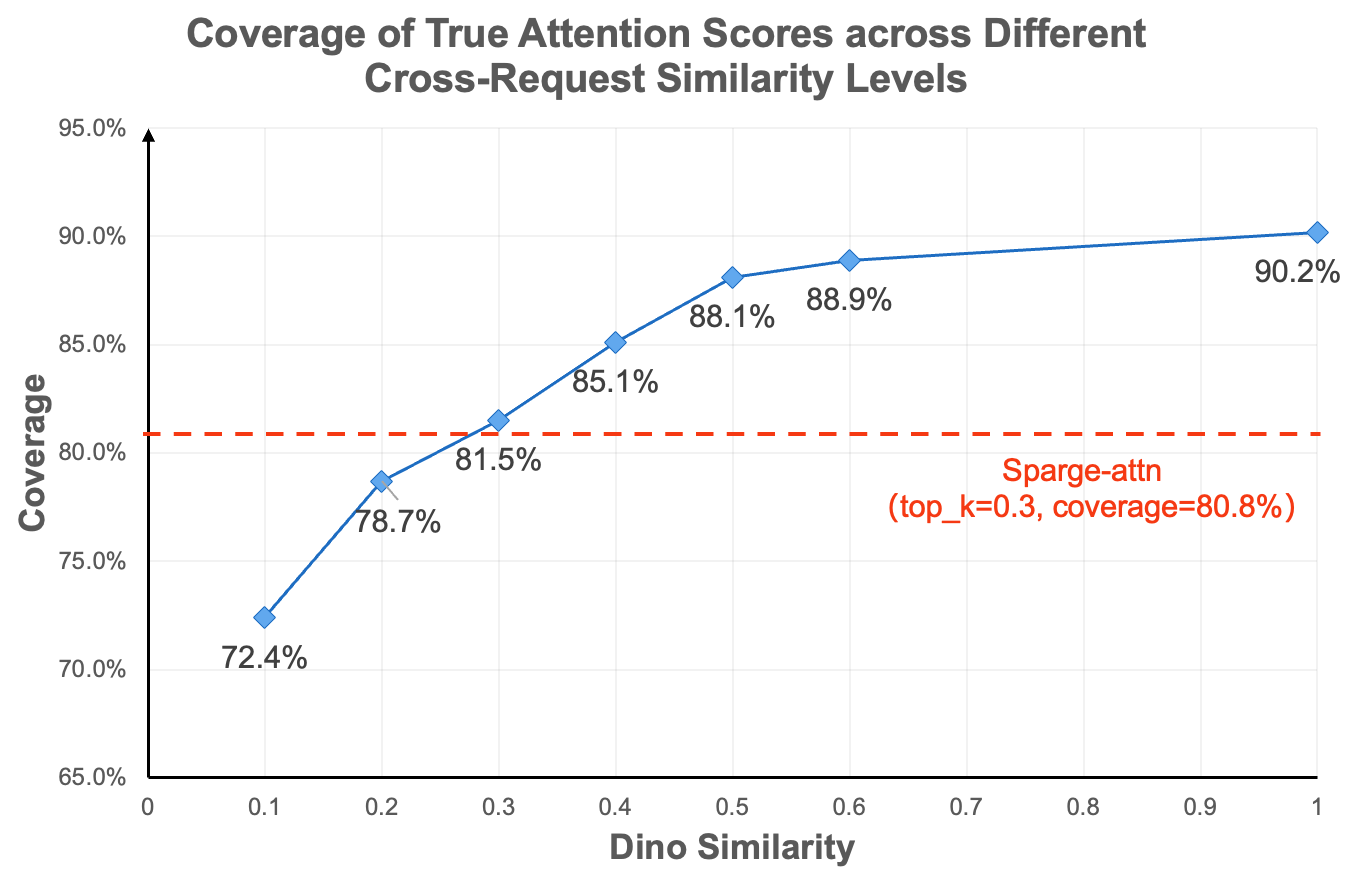}
\caption{GT attention coverage under different DINO similarities.}
\label{fig:coverage_with_diff_similarity}
\end{wrapfigure}

The zero-overhead coverage reported above is not unconditional; it hinges on how semantically close the retrieved request is to the target. To make this dependence explicit, we plot the GT attention coverage under the top-$p=0.9$ setting against the cross-request semantic similarity (measured by DINO) in Fig.~\ref{fig:coverage_with_diff_similarity}.
We observe a strong, monotonic positive correlation: as the DINO similarity between the retrieved source request and the current target request increases, the retrieved mask's coverage of the target's true attention mass rises significantly.

Concretely, once the semantic similarity exceeds \textbf{0.3}, the retrieved mask already attains higher coverage than the coarse-grained online prediction of Sparge-Attn, while incurring virtually no routing overhead.
This quantitative finding further validates the efficacy of cross-request sparsity reuse and justifies our database-driven retrieval threshold $\tau$: by filtering out dissimilar requests and falling back to dense attention when similarity is low, we can guarantee high fidelity in our sparse attention execution.

\subsection{Ablation on Downsampled Latent Feature Reuse and Guidance Enhancement}
\label{sec:ablation_bc}

This ablation jointly evaluates the two stabilization mechanisms used by the optional feature-reuse path: downsampled context computation, which preserves a globally coherent receptive field to suppress the boundary tearing caused by region-isolated reuse, and guidance enhancement, which recalibrates the image condition weakened by reuse.
The two are complementary and should jointly improve visual quality and condition alignment at negligible extra latency.

\begin{table}[t]
\centering
\caption{Ablation study on the feature-reuse-only path (downsampled latent feature reuse and guidance enhancement), \emph{without} sparsity reuse.}
\label{tab:ablation_downsample_enhance}
\small
\setlength{\tabcolsep}{3pt}
\renewcommand{\arraystretch}{1.08}
\resizebox{\linewidth}{!}{
\begin{tabular}{lccccccc}
\toprule
\textbf{Method} & \textbf{Downsample} & \textbf{Enhance} & \textbf{VBench-Q} $\uparrow$ & \textbf{DINO-I} $\uparrow$ & \textbf{CLIP-T} $\uparrow$ & \textbf{Latency} $\downarrow$ & \textbf{Speedup} $\uparrow$ \\
\midrule
SageAttention (baseline)            & --  & --  & 0.789 & 0.922 & 24.05 & 91.8s & 1.00$\times$ \\
Feature reuse w/o downsampling, w/o enhance  & No  & No  & 0.775 & 0.918 & 23.99 & 78.5s & 1.17$\times$ \\
Feature reuse w/o downsampling, w/  enhance  & No  & Yes & 0.777 & 0.921 & 24.03 & 78.6s & 1.17$\times$ \\
Feature reuse w/  downsampling, w/o enhance  & Yes & No  & 0.788 & 0.917 & 24.01 & 79.4s & 1.15$\times$ \\
Feature reuse w/  downsampling, w/  enhance  & Yes & Yes & 0.788 & 0.922 & 24.32 & 79.5s & 1.15$\times$ \\
\bottomrule
\vspace{-5pt}
\end{tabular}}
\vspace{-15pt}
\end{table}

Downsampling consistently improves visual quality, lifting VBench-Q to $0.788$ in both settings (from $0.775$ without and $0.777$ with enhancement), confirming that the coherent receptive field it preserves yields a more stable reuse signal beyond merely lowering reuse cost.
Guidance enhancement mainly strengthens condition consistency, improving DINO-I ($0.917\!\to\!0.922$) and CLIP-T ($24.01\!\to\!24.32$) under downsampling, thereby compensating for the image-condition weakening introduced by reuse.
Combining both yields the best quality--efficiency trade-off, attaining the highest VBench-Q ($0.788$), DINO-I ($0.922$), and CLIP-T ($24.32$) while holding latency at $79.5$\,s ($1.15\times$), essentially unchanged from the other variants.

\section{Conclusion and Limitations}
This paper presents a cross-request reuse framework for efficient image-to-video (I2V) serving under few-step distilled video diffusion models.
Instead of relying only on redundancy across denoising steps within a single request, we exploit the redundancy shared by similar requests in real I2V workloads.
Our key observation is that similar I2V requests exhibit highly consistent sparse attention patterns, allowing high-quality sparse masks from historical requests to be reused as request-conditioned priors with nearly zero online routing overhead.
Based on this observation, sparsity reuse avoids expensive per-request sparse mask search while preserving request adaptivity and mask quality.
We further combine it with feature reuse based on downsampled context computation to exploit latent redundancy without boundary artifacts, and with lightweight guidance enhancement to mitigate condition weakening and semantic drift introduced by reuse.
Experiments on Wan2.2-I2V show that our default sparsity-reuse configuration preserves generation quality while achieving a \textbf{2.16$\times$} end-to-end speedup, matching SVG2-level fidelity to the dense baseline with roughly half the latency.
When combined with feature reuse, the framework further reaches a \textbf{2.59$\times$} speedup while maintaining condition-consistent generation.

The current framework also has several limitations.
Specifically, the acceleration efficacy of our framework inherently depends on the semantic similarity across incoming requests, as a lack of similar historical queries will trigger fallback to baseline inference, thereby diluting the speedup benefits.
Furthermore, while database initialization is remarkably fast in domain-specific scenarios (e.g., effect template generation), extending our framework to general workloads may necessitate constructing and maintaining a larger-scale retrieval database to ensure high cache hit rates.

\bibliographystyle{plain}
\bibliography{modified_references}

\appendix

\section{Head level schedule for load balance}

After sparsification, the per-head attention cost becomes highly uneven, so with a fixed head-to-rank assignment some GPUs are left far busier than others. As the sequence-parallel attention is synchronized at every layer, the end-to-end latency is dictated by the slowest GPU, and this imbalance directly wastes compute. Crucially, the sparsity mask is known ahead of the attention computation, so each head's cost can be estimated in advance. We exploit this by running a greedy algorithm that re-partitions (reorders) heads across ranks \emph{before} the all-to-all communication, equalizing the per-GPU load without introducing any extra communication. In practice this lowers the per-layer load imbalance (max/avg over GPUs) from about $1.34$ to $1.12$.

Table~\ref{tab:head_level} reports the resulting end-to-end latency. On a single GPU the schedule is a no-op (there is nothing to balance across ranks), so the two rows match, and the benefit grows with the GPU count as the head assignment becomes increasingly skewed. At 8 GPUs the reordering reduces latency from $16.2$\,s to $15.1$\,s, a $\sim$7.28\% speedup, without any loss in generation quality.

\begin{table}[H]
\centering
\caption{Effect of head-level scheduling for load balance under different GPU counts. We report the end-to-end inference latency (seconds; lower is better).}
\label{tab:head_level}
\small
\setlength{\tabcolsep}{4.5pt}
\renewcommand{\arraystretch}{1.15}
\begin{tabular}{lcccc}
\toprule
\textbf{Schedule} & \textbf{1 GPU} & \textbf{2 GPUs} & \textbf{4 GPUs} & \textbf{8 GPUs} \\
\midrule
w/o scheduler  & 78.9   &    45.8   &     29.9  &  16.2 \\
w/ scheduler & 78.9   &    45.6   &   29.0 & \textbf{15.1}  \\
\bottomrule
\end{tabular}
\end{table}

\end{document}